\documentclass[letterpaper, 10 pt, conference]{ieeeconf}  

\usepackage[T1]{fontenc}
\usepackage[utf8]{inputenc}
\usepackage{authblk}
\usepackage{graphicx}
\usepackage[spaces,hyphens]{url}
\usepackage{algorithm}
\usepackage{algpseudocode}
\usepackage{algorithmicx}
\usepackage{array}
\usepackage{mathtools}
\usepackage{subcaption}
\usepackage{amssymb}
\usepackage{amsmath}
\usepackage{tikz}
\usepackage{diagbox}
\usepackage{flushend}
\usepackage{upgreek}
\usepackage{bm}
\usepackage{array}
\DeclareMathAlphabet{\mathcal}{OMS}{cmsy}{m}{n}
\usepackage{makecell}
\usepackage{stackengine}
\usepackage{multirow}
\usepackage[symbol]{footmisc}
\usepackage{sidecap}
\usepackage{tablefootnote}
\usepackage[flushleft]{threeparttable}
\usepackage{booktabs,caption}
\usepackage{xcolor, soul}
\usepackage{hyperref}
\sethlcolor{yellow}

\IEEEoverridecommandlockouts                              

\title{\LARGE \bf
Perception-Control Coupled Visual Servoing for Textureless Objects Using Keypoint-Based EKF
}

\author{
Allen Tao\textsuperscript{1,2*}, Jun Yang\textsuperscript{1*$\dagger$}, Stanko Oparnica\textsuperscript{1}, and Wenjie Xue\textsuperscript{1}
\thanks{*Equal contribution.}
\thanks{$^\dagger$Project lead and corresponding author.}
\thanks{\textsuperscript{1}The authors are with Epson Canada Ltd., Toronto, Canada.
        {\tt\footnotesize \{Jun.Yang, Stanko.Oparnica, Mark.Xue\}@ea.epson.com}}
\thanks{\textsuperscript{2}The author is with the University of Toronto, Toronto, Canada. This work was done during an internship at Epson Canada Ltd.
        {\tt\footnotesize allen.tao@robotics.utias.utoronto.ca}}
\thanks{This work was supported by Epson Canada Ltd.}
}

\begin{document}

\maketitle
\thispagestyle{empty}
\pagestyle{empty}

\begin{abstract}
Visual servoing is fundamental to robotic applications, enabling precise positioning and control. However, applying it to textureless objects remains a challenge due to the absence of reliable visual features. Moreover, adverse visual conditions, such as occlusions, often corrupt visual feedback, leading to reduced accuracy and instability in visual servoing. In this work, we build upon learning-based keypoint detection for textureless objects and propose a method that enhances robustness by tightly integrating perception and control in a closed loop. Specifically, we employ an Extended Kalman Filter (EKF) that integrates per-frame keypoint measurements to estimate 6D object pose, which drives pose-based visual servoing (PBVS) for control. The resulting camera motion, in turn, enhances the tracking of subsequent keypoints, effectively closing the perception-control loop. Additionally, unlike standard PBVS, we propose a probabilistic control law that computes both camera velocity and its associated uncertainty, enabling uncertainty-aware control for safe and reliable operation. We validate our approach on real-world robotic platforms using quantitative metrics and grasping experiments, demonstrating that our method outperforms traditional visual servoing techniques in both accuracy and practical application.

\end{abstract}

\section{INTRODUCTION}
Visual servoing (VS) refers to the use of visual feedback to control robot motion and serves as a key technique in high-precision tasks. Common approaches, such as image-based visual servo (IBVS), pose-based visual servo (PBVS) and hybrid approaches~\cite{chaumette2006visual}, rely on accurate 2D–2D or 2D–3D keypoint correspondences to estimate the image Jacobian or camera pose for control. Classical VS approaches typically use hand-crafted features~\cite{lowe1999object} to extract these correspondences. However, textureless objects, which are commonly found in industrial environments, pose significant challenges. The absence of distinctive visual features makes establishing reliable correspondences difficult, ultimately degrading servoing performance.

With advances in deep learning, many methods have been developed to reduce classical visual servoing’s reliance on hand-crafted features~\cite{saxena2017exploring, bateux2018training, yu2019siamese, adrian2022dfbvs, luo2024uncalibrated, chen2024cns}. Early works~\cite{saxena2017exploring, bateux2018training} employ convolutional neural networks (CNNs) to estimate relative camera pose and integrate this into a PBVS framework. However, these approaches are typically trained in specific scenes and often struggle to generalize to unseen environments. More recent approaches use CNNs~\cite{adrian2022dfbvs, luo2024uncalibrated} or Transformer-based networks~\cite{scherl2025vit, chen2024cns} to improve the reliability of 2D correspondence extraction on textureless objects, and incorporate these features into an IBVS controller. Despite these advances, IBVS suffers from inherent limitations, such as susceptibility to local minima and a limited convergence basin.

\begin{figure}[t]
\centering
  \includegraphics[width=0.9\linewidth]{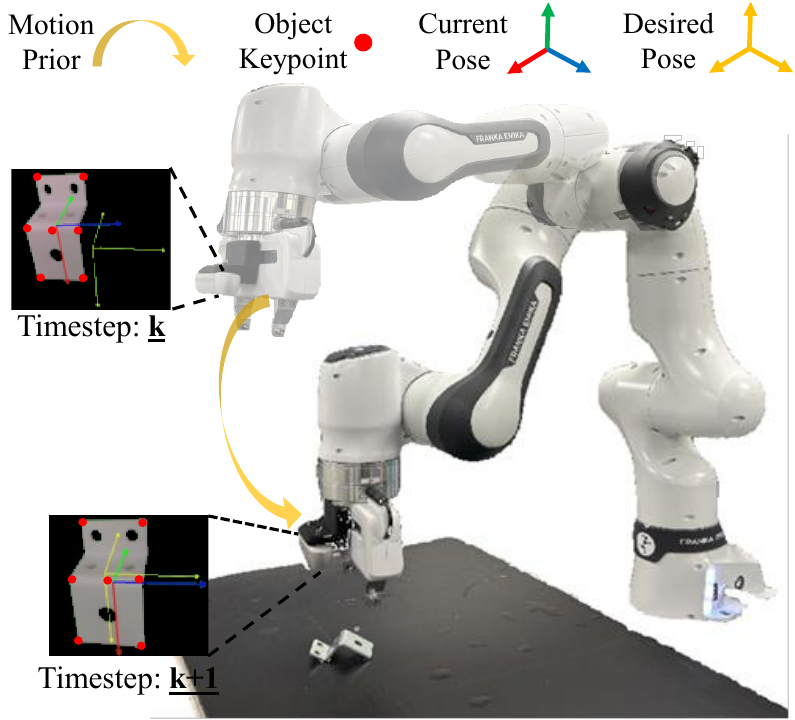}
\caption{Perception–control coupled visual servoing framework. An EKF integrates complementary information from keypoints and the motion prior, producing reliable 6D object poses for visual servoing.}
\label{fig_vis}
\end{figure}

On the other hand, recent progress in RGB-based object pose estimation~\cite{xiang2018posecnn, li2018deepim, park2019pix2pose, peng2019pvnet, labbe2022megapose, xu20246d, yang2025active} provides a promising alternative for enabling PBVS on textureless objects. These methods assume access to a known 3D object model and estimate the 6D object pose from a single RGB image. Although they can be seamlessly integrated into PBVS frameworks, they typically decouple perception from control and rely solely on single-frame predictions, thereby neglecting temporal information. As a result, they often fail under adverse visual conditions, such as occlusions or abrupt illumination changes, leading to unstable control and reduced robustness in visual servoing. Moreover, most pose estimators~\cite{xiang2018posecnn, li2018deepim, labbe2022megapose, xu20246d} produce only a single deterministic pose estimate, which results in fixed control velocities that do not account for uncertainty. To ensure safe and robust robotic operation, it is crucial to model these uncertainties and leverage them to regulate control.

To overcome these limitations, we propose a perception–control coupled framework that leverages temporal information to enable robust visual servoing of textureless objects and maintain performance under adverse conditions. As illustrated in Figure~\ref{fig_vis}, the framework employs an Extended Kalman Filter (EKF) to integrate complementary information from keypoint observations and the motion prior, producing reliable 6D object poses for pose-based visual servoing (PBVS). Unlike many EKF-based trackers that rely on additional sensors such as IMU~\cite{weiss2011real} or incorporate velocity into the state vector~\cite{mehralian2020ekfpnp, lin2022keypoint}, our method derives its motion prior directly from the camera velocity computed within the visual servoing loop. 

Figure~\ref{pipeline} shows the key steps of our framework. We first employ PVNet~\cite{peng2019pvnet}, a state-of-the-art approach, to acquire per-frame object keypoints. Our framework then operates in a closed-loop perception–control cycle: during perception, we use the EKF to integrate keypoints and the motion prior to output a reliable 6D pose; during control, the estimated pose is fed into a probabilistic control law that outputs both the camera velocity and its associated uncertainty; finally, the predicted velocity is used to actuate the robot and propagated into the next perception stage for state prediction. Importantly, the velocity uncertainty computed in the control stage can be leveraged to enhance operational safety, e.g., for collision avoidance.

In summary, our key contributions are:
\begin{itemize}
    \item We propose a perception-control coupled visual servoing framework designed for textureless objects. By leveraging temporal information, our approach enhances PBVS accuracy while preserving robust performance under challenging conditions, such as occlusions or lighting changes.
    \item To account for measurement uncertainties, we formulate a probabilistic control law that incorporates this uncertainty into the computed velocity commands, enhancing the safety of robotic operations.
\end{itemize}

\section{RELATED WORK}
\label{related work}
\subsection{Visual Servoing}
Visual servoing (VS) is a control strategy that guides a robot’s motion using visual feedback. It is commonly divided into two types based on how the error is defined~\cite{chaumette2006visual}. In IBVS, the error is defined in image space, while in PBVS, it is defined in Euclidean space. IBVS computes the control velocity either from keypoint correspondences~\cite{feddema2002vision,tran2007real}, or directly from the pixel intensities of the entire image~\cite{collewet2011photometric, marchand2019subspace, felton2022visual}. PBVS utilizes the relative pose between current and desired camera pose to compute the velocity control command~\cite{thuilot2002position}. Generally, PBVS offers a larger convergence region and more efficient 3D trajectories than IBVS, but it is more computationally demanding due to the need for 6D pose estimation at each frame.

Recently, VS has increasingly adopted deep learning techniques to enhance performance. Some studies use deep learning to improve keypoint estimation, thereby facilitating IBVS~\cite{adrian2022dfbvs, luo2024uncalibrated, chen2024cns, scherl2025vit}. For PBVS, methods such as~\cite{saxena2017exploring, bateux2018training} use CNNs to estimate the relative camera pose, which is then integrated into a PBVS framework. Another line of work replaces the velocity control law with neural networks~\cite{yu2023hyper, yu2019siamese, puang2020kovis, felton2021siame}, but sacrifices theoretical guarantees like convergence and trajectory optimality in PBVS. In contrast, our approach builds upon the PBVS formulation while incorporating deep learning techniques to improve performance without discarding these guarantees.

\begin{figure*}[t]
\centering
  \includegraphics[width=0.98\linewidth]{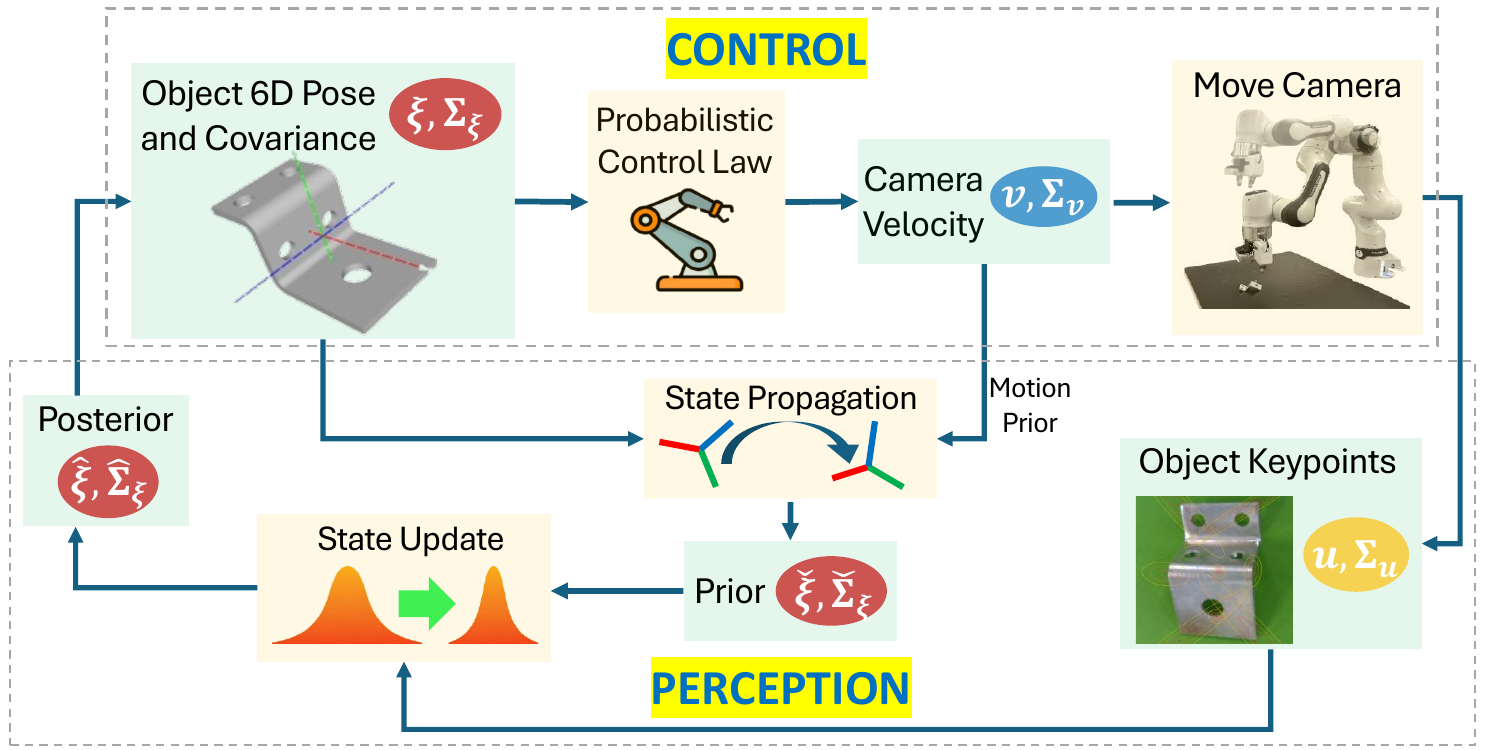}
\caption{Our framework operates in a closed-loop cycle: during the perception stage, the EKF fuses keypoints and the motion prior to estimate 6D poses; during the control stage, a probabilistic control law computes camera velocity, which actuates the robot and serves as the motion prior for the next perception stage.}
\label{pipeline}
\end{figure*}

\subsection{Object Pose Estimation}
Advances in object pose estimation have made visual servoing for textureless objects increasingly feasible. Given a known 3D model, RGB-based pose estimators~\cite{xiang2018posecnn, park2019pix2pose, peng2019pvnet, li2018deepim, labbe2022megapose} can recover the object's 6D pose, which can then be directly incorporated into a PBVS framework. Leveraging CNNs or Vision Transformers (ViTs), these approaches either regress the 6D pose directly from a single RGB image~\cite{xiang2018posecnn, li2018deepim, labbe2022megapose} or estimate it by first predicting 2D–3D correspondences followed by a PnP algorithm~\cite{park2019pix2pose, peng2019pvnet}.

While single-view pose estimators can be effective, they often fail in the presence of occlusions or lighting variations. To improve robustness, recent works incorporate temporal information for pose estimation~\cite{labbe2020cosypose,deng2021poserbpf,lin2022keypoint,merrill2022symmetry,yang20236d}. These methods either assume known camera poses~\cite{labbe2020cosypose,yang20236d,yang2025active}, or jointly estimate both camera and object poses, a strategy known as object-level SLAM~\cite{deng2021poserbpf,lin2022keypoint,merrill2022symmetry}. Temporal information enables more consistent and accurate pose estimates, especially in challenging conditions such as occlusion, motion blur, or weak textures. In this work, we leverage temporal cues through a keypoint-based EKF to estimate robust 6D object poses, enabling reliable visual servoing.

\section{FRAMEWORK OVERVIEW}
The proposed framework, illustrated in Figure~\ref{pipeline}, operates in a closed-loop fashion, executing a cycle of perception and control at every frame. At the perception stage (Section~\ref{perception}), we aim to estimate the 6-DOF object pose, $\boldsymbol{T}_{co}$, which represents the transformation from the object frame $\mathcal{F}_o$ to the camera frame $\mathcal{F}_c$:
\begin{equation}
\label{equ_pose}
\boldsymbol{T}_{co} =
\begin{bmatrix}
\mathbf{C}_{co} & \mathbf{t}_{co} \\
\mathbf{0}^\top & 1
\end{bmatrix} \in \mathbb{SE}(3),
\end{equation}
where $\mathbf{C}_{co} \in \mathbb{SO}(3)$ denotes the object orientation with respect to the camera frame, and $\mathbf{t}_{co} \in \mathbb{R}^3$ represents the object position. This pose estimate $\boldsymbol{T}_{co}$ is then passed to the control stage (Section~\ref{control}), which computes the camera velocity $\mathbf{v}_c$: 
\begin{equation}
    \mathbf{v}_c =
    {\left[
    \mathbf{v}_p \;\;\; \boldsymbol{\omega}
    \right]}\in\mathbb{R}^{6},
\end{equation}
with $\mathbf{v}_p\in\mathbb{R}^{3}$ as the translational velocity and $\boldsymbol{\omega}\in\mathbb{R}^{3}$ as the angular velocity, both expressed in the camera frame $\mathcal{F}_{c}$. Finally, this estimated velocity is fed back as the motion prior for the next perception stage, completing the closed loop.

\section{PERCEPTION}
The goal of the perception stage is to estimate the 6-DOF object pose, $\boldsymbol{T}_{co}$. We achieve this using a keypoint-based Extended Kalman Filter (EKF), which fuses the motion prior from the previous frame with keypoint measurements from the current frame. The EKF then performs state prediction (Section~\ref{prediction}) and update (Section~\ref{update}) to produce a refined pose estimate.

\label{perception}
\subsection{State Representation and Propagation}
\label{prediction}
\textbf{State Propagation}. We represent the EKF state vector, $\mathbf{x}$, using the 6-DOF object pose:
\begin{equation}
    \mathbf{x} = \left[\mathbf{t}_{co} \;,\; \mathbf{C}_{co}\right],
\end{equation}
where $\mathbf{t}_{co}$ and $\mathbf{C}_{co}$ are defined in Equation~\ref{equ_pose}. For brevity, we will omit the subscript “${co}$” in the following sections. To predict the state at the next timestep, $k\!+\!1$, we propagate the current state, $\mathbf{x}_k$, using a discrete-time constant-velocity motion model, incorporating the camera's translational velocity $\mathbf{v}_{p,k}$ and angular velocity $\boldsymbol{\omega}_k$:
\begin{equation}
\label{propagate_trans}
    \check{\mathbf{t}}_{k+1} = \exp{\left(-\boldsymbol{\omega}_k^{\wedge}\Delta t\right)} \mathbf{t}_{k} - \mathbf{v}_{p,k}\Delta t,
\end{equation}
\begin{equation}
\label{propagate_rot}
    \check{\mathbf{C}}_{k+1} = \exp{\left(-\boldsymbol{\omega}_k^{\wedge}\Delta t\right)} \mathbf{C}_{k},
\end{equation}
where $\check{\left(\cdot\right)}$ denotes propagated quantities. $\Delta t$  is the time interval between consecutive time steps, and ${\wedge}$ represents the skew-symmetric operator applied to a vector. The acquisition of the camera's velocity $\mathbf{v}_{c,k} = \left[\mathbf{v}_{p,k} ,\; \boldsymbol{\omega}_k\right]$ at timestep $k$ will be described in Section~\ref{control}.

\textbf{Error State Propagation}. Alongside the state mean, the EKF must also propagate the associated uncertainty. To achieve this, we define an error state in which the rotational component is linearized using the Lie algebra $\mathfrak{so}(3)$. The resulting 6-dimensional error state vector, $\delta\mathbf{x}$, is defined as:
\begin{equation}
    \delta\mathbf{x} = \left[\delta\mathbf{t} \;,\; \delta\boldsymbol{\phi}\right],
\end{equation}
where $\delta \mathbf{t} \in \mathbb{R}^3$ and $\delta \boldsymbol{\phi} \in \mathbb{R}^3$ represent the translational and rotational errors, respectively.

By linearizing the motion model in Equations~(\ref{propagate_trans}) and (\ref{propagate_rot}) around the current state estimate, the discrete-time error-state propagation is given by:
\begin{equation}
\label{propagate_error}
    {\begin{bmatrix}
    \delta\check{\mathbf{t}}_{k+1} \\
    \delta\check{\boldsymbol{\phi}}_{k+1}
    \end{bmatrix}}= 
    \mathbf{F}
    {\begin{bmatrix}
    \delta\mathbf{t}_{k} \\
    \delta\boldsymbol{\phi}_{k}
    \end{bmatrix}}
    + \mathbf{G}\mathbf{n},
\end{equation}
where $\mathbf{n} = {\left[ \mathbf{n_t}^T \;\;\mathbf{n_{\boldsymbol{\phi}}}^T \right]}^T$ represents the camera velocity noise, and $\mathbf{G}$ is the noise Jacobian. The error-state transition matrix, $\mathbf{F}_k$, is:
\begin{equation}
    \mathbf{F}_k = 
    {\begin{bmatrix}
    \frac{\partial \exp{\left(-\omega_k^{\wedge}\Delta t\right)}\mathbf{t}_k}{\partial \mathbf{t}_k} & \mathbf{0}_{3\times3} \\
    \mathbf{0}_{3\times3} & \frac{\partial \text{ln}{\left(\exp{(-\boldsymbol{\omega}_k^{\wedge}\Delta t)} \mathbf{C}_{k}\right)}^{\vee}}{\partial \boldsymbol{\phi}_k}
    \end{bmatrix}},
\end{equation}
with:
\begin{equation}
    \frac{\partial \exp{\left(-\boldsymbol{\omega}_k^{\wedge}\Delta t\right)}\mathbf{t}_k}{\partial \mathbf{t}_k} = \exp{\left(-\boldsymbol{\omega}_k^{\wedge}\Delta t\right)},
\end{equation}
\begin{equation}
\label{Jac_rot}
\frac{\partial \text{ln}{\left(\exp{(-\boldsymbol{\omega}_k^{\wedge}\Delta t)} \mathbf{C}_{k}\right)}^{\vee}}{\partial \boldsymbol{\phi}_k}=\mathcal{J}_r^{-1}\left(\text{ln}{\left(\exp{(-\boldsymbol{\omega}_k^{\wedge}\Delta t)} \mathbf{C}_{k}\right)}^{\vee}\right).
\end{equation}
Here, $\text{ln}\left(\cdot\right)^{\vee}$ denotes the logarithmic mapping from Lie group $\mathbb{SO}(3)$ to its Lie algebra $\mathfrak{so}(3)$. The operator $\mathcal{J}_r(\cdot)$ is the right Jacobian. This Jacobian is derived using the right perturbation model together with the Baker-Campbell-Hausdorff (BCH) approximation. For detailed derivations, readers are referred to~\cite{barfoot2024state}.

Finally, the state covariance $\mathbf{P}$ is propagated to the next timestep, $k+1$, as follows:
\begin{equation}
\label{cov_velocity}
\mathbf{R} = \mathsf{diag}\left(\sigma_{v_p} \mathbf{I} \;,\;\sigma_{{v_\omega}} \mathbf{I}\right),
\end{equation}
\begin{equation}
\label{cov_propagate}
\check{\mathbf{P}}_{k+1}=\mathbf{F}_{k} \mathbf{P}_k\mathbf{F}_{k}^T + \mathbf{G}_{k} \mathbf{R}\mathbf{G}^T_k \Delta t,
\end{equation}
where $\sigma_{v_p}$ and $\sigma_{{v_\omega}}$ denote the standard deviations of the translational and angular components of the camera velocity noise, respectively, and $\mathbf{G}_{k}$ is the identity noise Jacobian.

\subsection{Measurement Model and Update}
\label{update}
While the propagation step provides a prior estimate of the pose, it is susceptible to drift over time. To correct this, the proposed EKF performs measurement updates using 2D–3D keypoint correspondences obtained from per-frame RGB images.
\subsubsection{\textbf{Keypoint Measurements}}
\label{kps}
\begin{figure}[t]
\centering
\begin{subfigure}{0.23\textwidth}
  \includegraphics[width=\linewidth]{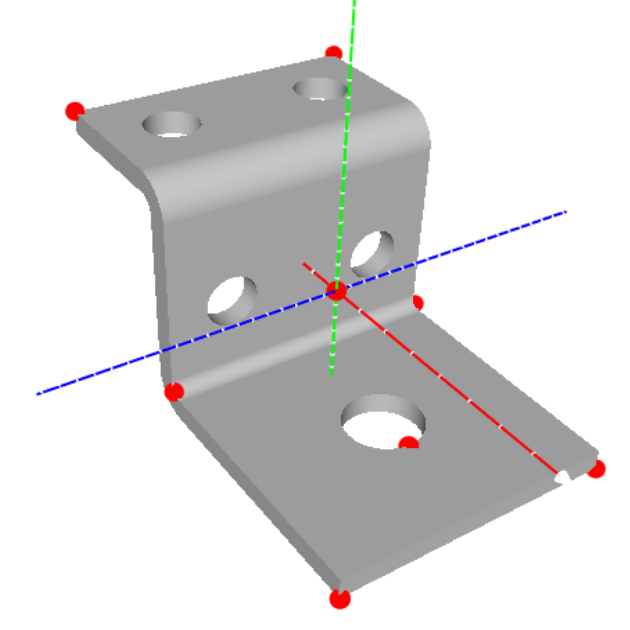}
\end{subfigure}
\begin{subfigure}{0.23\textwidth}
    \includegraphics[width=\linewidth]{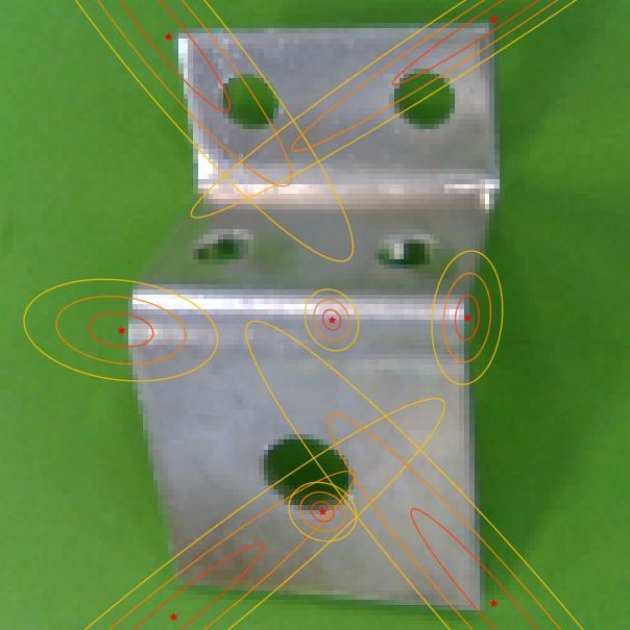}         
\end{subfigure}
\caption{Example of 3D object points $\mathbf{X}_o$, along with 2D image keypoints, $\mathbf{u}_i$, and their associated uncertainties, $\boldsymbol{\Sigma}_{\mathbf{u}_i}$, as estimated using PVNet~\cite{peng2019pvnet}.}
\label{fig_kps}
\end{figure}

In the update step, the measurement comprises 2D image keypoints associated with 3D object points $\mathbf{X}_o$. For each object, we first select $N$ 3D keypoints on the CAD model using farthest point sampling (FPS). At runtime, the corresponding 2D keypoints are estimated using PVNet~\cite{peng2019pvnet}, which predicts their image locations and provides an associated covariance matrix for each keypoint to model localization uncertainty. For timestep~$k\!+\!1$, the measured keypoints and their uncertainties are stacked into the measurement vector $\mathbf{z}_{k+1}$ and covariance matrix $\mathbf{Q}_{k+1}$, defined as:
\begin{align}
    \mathbf{z}_{k+1} =
    \begin{bmatrix}    
    \mathbf{u}_{0} \\    
    \vdots \\
    \mathbf{u}_{N-1}
    \end{bmatrix}
    \:\:,\:\:
    \mathbf{Q}_{k+1} =
    \begin{bmatrix}
    \boldsymbol{\Sigma}_{\mathbf{u}_{0}} & {} & {}\\
    {} & \ddots & {}\\
    {} & {} & \boldsymbol{\Sigma}_{\mathbf{u}_{N-1}}\\
    \end{bmatrix},
\end{align}
where $\mathbf{u}_{i} \in \mathbb{R}^2$ denotes the \emph{measured} 2D keypoint of the $i$-th object point and $\boldsymbol{\Sigma}_{\mathbf{u}_{i}} \in \mathbb{R}^{2\times 2}$ its covariance. Figure~\ref{fig_kps} shows an example of the 3D object points $\mathbf{X}_o$, along with the measured keypoints, $\mathbf{u}_i$, and their associated covariance matrices, $\boldsymbol{\Sigma}_{\mathbf{u}_i}$. Note that while we adopt PVNet, other keypoint detection methods~\cite{park2019pix2pose,merrill2022symmetry} can also be integrated into our EKF framework.

\subsubsection{\textbf{Measurement Model}}
We define the measurement model $h(\cdot)$ by first transforming each 3D object point $\mathbf{X}_{o,i} \in \mathbb{R}^{3}$ from the object frame $\mathcal{F}_o$ to the camera frame $\mathcal{F}_c$ using the object pose estimate $\left(\check{\mathbf{C}}_{k+1}, \check{\mathbf{t}}_{k+1}\right)$, and then projecting the result onto the image plane:
\begin{align}
\label{measurement model}
\check{\mathbf{u}}_{i} &= h\left(\check{\mathbf{C}}_{k+1}, \check{\mathbf{t}}_{k+1}\right) \\
&= \text{proj}\!\left( \check{\mathbf{C}}_{k+1}\mathbf{X}_{o, i} + \check{\mathbf{t}}_{k+1} \right)
\end{align}
where $\check{\mathbf{u}}_i \in \mathbb{R}^2$ is the \emph{predicted} 2D location of the $i$-th object point, and $\text{proj}(\cdot)$ denotes the perspective projection function. The residual error $\boldsymbol{\epsilon}_{k+1}$ at frame~$k\!+\!1$ is defined as the difference between the measured and predicted keypoints:
\begin{equation}
\label{residual}
\boldsymbol{\epsilon}_{k+1} = \mathbf{z}_{k+1} - \check{\mathbf{z}}_{k+1} = 
    \begin{bmatrix}    
    \mathbf{u}_{0} - \check{\mathbf{u}}_{0}\\    
    \vdots \\
    \mathbf{u}_{N-1} - \check{\mathbf{u}}_{N-1}
    \end{bmatrix},
\end{equation}
where $\mathbf{z}_{k+1} = [\mathbf{u}_{0}^\top, \dots, \mathbf{u}_{N-1}^\top]^\top$ is the vector of measured keypoints from PVNet, and $\check{\mathbf{z}} = [\check{\mathbf{u}}_{0}^\top, \dots, \check{\mathbf{u}}_{N-1}^\top]^\top$ is the corresponding vector of predicted keypoints.

To acquire the measurement Jacobian $\mathbf{H}_i$ for each keypoint residual $\boldsymbol{\epsilon}_i$, we differentiate the residual $\boldsymbol{\epsilon}_i$ with respect to the error state $\delta\mathbf{x}$:
\begin{equation}
\mathbf{H}_i = \frac{\partial \boldsymbol{\epsilon}_i}{\partial \delta\mathbf{x}} = 
{\begin{bmatrix} 
\frac{\partial \boldsymbol{\epsilon}_i}{\partial \delta\mathbf{t}_{k+1}} \\[0.5em]
\frac{\partial \boldsymbol{\epsilon}_i}{\partial \delta\boldsymbol{\phi}_{k+1}}
\end{bmatrix}},
\end{equation}
where
\begin{equation}
\frac{\partial \boldsymbol{\epsilon}_i}{\partial \delta\mathbf{t}_{k+1}}
=\frac{\partial \left(\mathbf{u}_{i} - \check{\mathbf{u}}_{i}\right)}{\partial \delta\mathbf{t}_{k+1}}
=-\frac{\partial h\left(\mathbf{C}_{k+1},\mathbf{t}_{k+1}\right)}{\partial \delta\mathbf{t}_{k+1}},
\end{equation}
\begin{equation}
\label{Derivative_Rot}
\frac{\partial \boldsymbol{\epsilon}_i}{\partial \delta\boldsymbol{\phi}_{k+1}}
=\frac{\partial \left(\mathbf{u}_{i} - \check{\mathbf{u}}_{i}\right)}{\partial \delta\boldsymbol{\phi}_{k+1}}
=-\frac{\partial h\left(\mathbf{C}_{k+1},\mathbf{t}_{k+1}\right)}{\partial \delta\boldsymbol{\phi}_{k+1}}.
\end{equation}
To derive $\frac{\partial h\left(\mathbf{C}_{k+1},\mathbf{t}_{k+1}\right)}{\partial \delta\boldsymbol{\phi}_{k+1}}$ in Equation~\ref{Derivative_Rot}, we employ a left perturbation model in the Lie algebra $\mathfrak{so}(3)$, which represents the rotational update as a local increment on the manifold and enables its linearization for differentiation. The overall measurement Jacobian $\mathbf{H}_{k+1}$ at frame~$k\!+\!1$ is finally formed by stacking the individual Jacobians $\mathbf{H}_{i,k+1}$ for all keypoints:
\begin{equation}
\mathbf{H}_{k+1} = {\left[\mathbf{H}_0, \mathbf{H}_1, \cdots, \mathbf{H}_{N-1}\right]}^T
\end{equation}

\subsubsection{\textbf{EKF Update}}
Once we obtained the measurement Jacobian $\mathbf{H}_{k+1}$, we can update our estimate using a regular EKF update:
\begin{equation}
\begin{aligned}
\mathbf{K}_{k+1} &= \check{\mathbf{P}}_{k+1}\mathbf{H}_{k+1}^T{\left(\mathbf{H}_{k+1}\check{\mathbf{P}}_{k+1}\mathbf{H}_{k+1}^T + \mathbf{Q}_{k+1}\right)}^{-1} \\
\hat{\mathbf{P}}_{k+1} &= \left(\mathbf{I} - \mathbf{K}_{k+1}\mathbf{H}_{k+1}\right)\check{\mathbf{P}}_{k+1}\\
\delta\mathbf{x}_{k+1} &= \mathbf{K}_{k+1}\boldsymbol{\epsilon}_{k+1}
\end{aligned}
\end{equation}
where $\delta\mathbf{x}_{k+1} = \left[\delta\mathbf{t}_{k+1} \;,\; \delta\boldsymbol{\phi}_{k+1}\right]$ is state update, to inject $\delta\mathbf{x}_{k+1}$ into the predicted nominal states, we apply:
\begin{equation}
\label{update_step}
\begin{aligned}
\hat{\mathbf{t}}_{k+1} &= \check{\mathbf{t}}_{k+1} + \delta\mathbf{t}_{k+1} \\
\hat{\mathbf{C}}_{k+1} &= \exp{\left(\delta\boldsymbol{\phi}_{k+1}^{\wedge}\right)}\check{\mathbf{C}}_{k+1} 
\end{aligned}
\end{equation}
where $\exp\left(\cdot\right)$ is the exponential mapping from the Lie algebra $\mathfrak{so}(3)$ to its Lie group $\mathbb{SO}(3)$, and $\hat{\left(\cdot\right)}$ denotes the updated quantities.

\section{PROBABILISTIC CONTROL}
\label{control}
For each frame $k$, given the updated EKF state $\hat{\mathbf{x}}$ (from Equation~\ref{update_step}), the objective of the control stage is to compute the camera velocity $\mathbf{v}_c$ from it. This velocity both actuates the robot and serves as the motion prior for state propagation in the next timestep (Equations~\ref{propagate_trans} and~\ref{propagate_rot}), thereby closing the perception–control loop. Additionally, to account for uncertainty in the estimated velocity, we estimate its covariance $\boldsymbol{\Sigma}_{\mathbf{v}_c}$, enabling uncertainty-aware motion execution.

To compute $\mathbf{v}_c$, we follow the classic PBVS control law, which derives the camera velocity through the desired and current object poses. Specifically, let $\mathbf{T}_{c^*c}$ denote the relative transformation from the current camera frame $\mathcal{F}_c$ to the desired camera frame $\mathcal{F}_{c^*}$, obtained as
\begin{equation}
\label{T_c*c}
\mathbf{T}_{c^*c} = 
\mathbf{T}_{c^*o} \cdot \mathbf{T}_{co}^{-1} =
{\begin{bmatrix} 
\mathbf{C}_{c^*c} & \mathbf{t}_{c^*c}\\
\mathbf{0}^T & 1
\end{bmatrix}}
\end{equation}
where $\mathbf{T}_{c^*o}$ represents the desired object pose, the current object pose $\mathbf{T}_{co}$ is assembled from the updated EKF states in Equation~\ref{update_step}:
\begin{equation}
\mathbf{T}_{co} = 
{\begin{bmatrix} 
\hat{\mathbf{C}}_{co} & \hat{\mathbf{t}}_{co}\\
\mathbf{0}^T & 1
\end{bmatrix}}
\end{equation}
then the mean of camera velocity $\mathbf{v}_c$ is computed using the PBVS control law~\cite{chaumette2006visual}:
\begin{equation}
\label{control_law}
\mathbf{v}_c = -\lambda
{\begin{bmatrix} 
\mathbf{C}_{c^*c}^T \:\: \mathbf{t}_{c^*c}\\
\theta \mathbf{u}
\end{bmatrix}}
\end{equation}
where ${\theta \mathbf{u}}$ is the axis–angle representation of the rotation matrix $\mathbf{C}_{c^*c}$, and $\lambda > 0$ is a control gain.

To acquire the velocity uncertainty, $\boldsymbol{\Sigma}_{\mathbf{v}_c}$, we linearize the system (Equations~\ref{T_c*c}-\ref{control_law}) and propagate the EKF state covariance $\hat{\mathbf{P}}$ through the PBVS control law. Specifically, we derive the Jacobian of the camera velocity with respect to the EKF error state vector, denoted as $\boldsymbol{\mathsf{J}}_{\delta\hat{\mathbf{x}}}$:
\begin{equation}
\boldsymbol{\mathsf{J}}_{\delta\hat{\mathbf{x}}} = 
\frac{\partial \mathbf{v}_c}{\partial \delta\hat{\mathbf{x}}} =
\frac{\partial \mathbf{v}_c}{\partial \boldsymbol{\xi}_{c*c}}
\frac{\partial \boldsymbol{\xi}_{c*c}}{\partial \delta\hat{\mathbf{x}}}
\end{equation}
where $\boldsymbol{\xi}_{c*c} \in \mathbb{R}^6$ is the Lie algebra representation of the relative pose $\mathbf{T}_{c^*c}$. The camera velocity covariance is then obtained via forward propagation:
\begin{equation}
\boldsymbol{\Sigma}_{\mathbf{v}_{c}} = \boldsymbol{\mathsf{J}}_{\delta\hat{\mathbf{x}}} \hat{\mathbf{P}} \boldsymbol{\mathsf{J}}_{\delta\hat{\mathbf{x}}}^T
\end{equation}
where $\boldsymbol{\Sigma}_{\mathbf{v}_{c}}$ is a $6 \times 6$ matrix. To quantify this uncertainty, we express it as the differential entropy:
\begin{align}
\label{equ_differential_entropy}
h_e\left( \boldsymbol{\Sigma}_{\mathbf{v}_{c}} \right) = \frac{1}{2} \ln{\left(\left(2\pi e\right)^n \left| \boldsymbol{\Sigma}_{\mathbf{v}_{c}} \right| \right)}
\end{align}
where $h_e\left( \boldsymbol{\Sigma}_{\mathbf{v}_{c}} \right)$ is expressed in nats. To incorporate the estimated velocity uncertainty into the robot control policy, we define an uncertainty threshold. When the entropy exceeds this threshold, the velocity is significantly reduced to ensure safety. This mechanism is crucial for maintaining safe robotic operation under uncertain motion estimates.

\section{EXPERIMENTS}
\subsection{Experiment Setup}
\label{setup}
\begin{figure}[t]
\centering
  \includegraphics[width=0.9\linewidth]{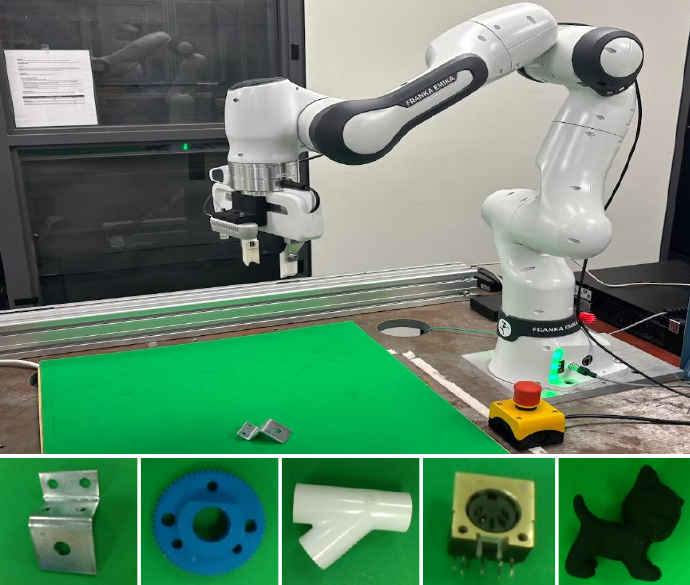}
\caption{Experimental setup. Top: Hardware platform for visual servoing. Bottom: Target objects for evaluation.}
\vspace{-1.0\baselineskip}
\label{fig_setup}
\end{figure}

We conduct experiments on a real-world robotic platform, shown in Figure~\ref{fig_setup}. The setup includes a 7-DoF Franka Emika arm with an Intel RealSense D435 camera mounted on the end-effector. A desktop with a 3.60 GHz CPU and NVIDIA A6000 GPU acts as the server. The Franka controller is built on the ViSP library~\cite{marchand2005visp}, communicating to the server via ZeroMQ~\cite{hintjens2013zeromq}. At each timestep, the server processes the camera image and sends velocity commands to the robot for real-time control.

We perform visual servoing experiments on five textureless objects with varied colors and materials (Figure~\ref{fig_setup}). In each episode, the desired camera pose $\mathbf{T}_{c^*o}$ is set roughly $15\;cm$ above the target, with $\pm5\;cm$ translational variation in XYZ. The initial pose $\mathbf{T}_{co}$ is sampled around $30\;cm$ above the object, with random translations of $\left[-10\;cm, 10\;cm\right]$ and rotations of $\left[-75^{\circ}, 75^{\circ}\right]$ applied. For each object, we run 30 visual servoing trials using our method and baseline approaches. As shown in Figure~\ref{fig_conditions}, trials are performed under diverse backgrounds, lighting conditions, and partial occlusions. To reflect difficulty, we group scenes into \textbf{Normal Conditions} (controlled settings) and \textbf{Adverse Conditions} (with challenging backgrounds, lighting, and occlusions).

\subsection{Implementation and Baselines}
\begin{figure}[t]
\centering
\begin{subfigure}{0.235\textwidth}
  \includegraphics[width=\linewidth]{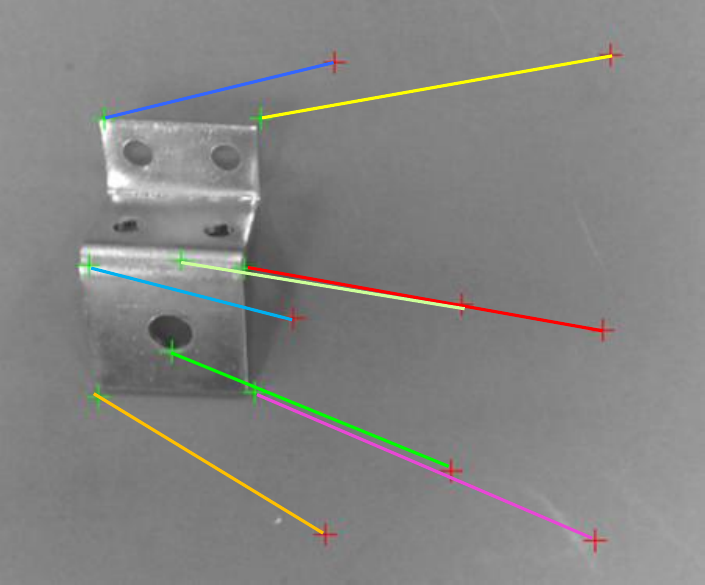}
  \caption{IBVS+PVNet}
  \label{fig_ibvs}
\end{subfigure}
\begin{subfigure}{0.235\textwidth}
    \includegraphics[width=\linewidth]{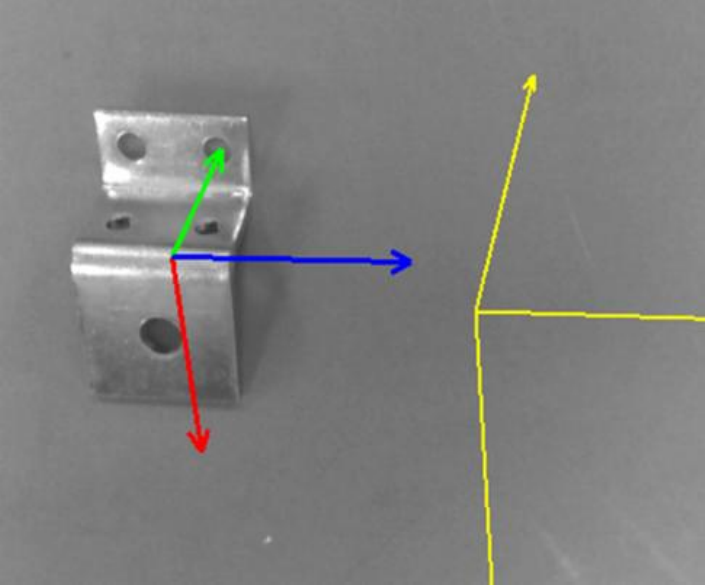}         
    \caption{PBVS+PVNet}
     \label{fig_pbvs}
\end{subfigure}
\begin{subfigure}{0.48\textwidth}
    \includegraphics[width=\linewidth]{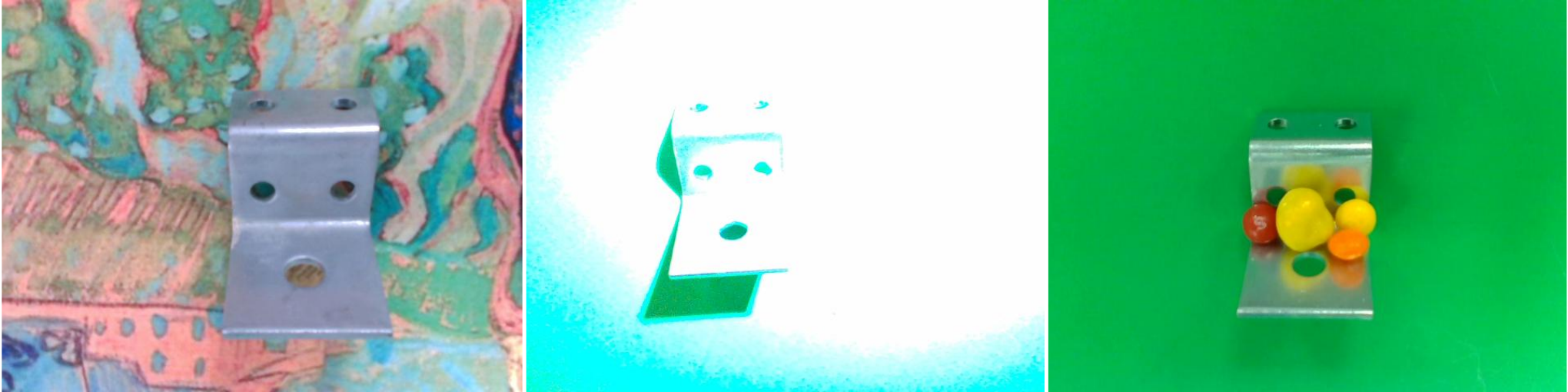}  
    \caption{Experiments with different visual backgrounds, lighting conditions, and partial occlusions.}
     \label{fig_conditions}
\end{subfigure}
\caption{Baseline approaches used in our evaluation. (a) IBVS+PVNet. (b) PBVS+PVNet. (c) Experiments with different backgrounds, lighting conditions, and occlusions.}
\label{fig_baselines}
\end{figure}

As discussed in Section~\ref{kps}, we use PVNet~\cite{peng2019pvnet} to extract 2D keypoints for textureless objects. For each object, PVNet is trained from scratch on approximately 200,000 synthetic images. We quantitatively evaluate our approach against classic IBVS and PBVS baselines. To ensure a fair comparison, we use the same trained PVNet model and integrate it into both baseline methods, referred to as IBVS-PVNet and PBVS-PVNet, respectively.
\begin{itemize}
    \item \textbf{IBVS+PVNet}. Figure~\ref{fig_ibvs} shows our integration of PVNet into the IBVS framework. We use the target camera pose $\mathbf{T}_{c^*o}$ to project 3D model points to obtain desired 2D keypoints $\mathbf{p}^*$. At each timestep, PVNet extracts current 2D keypoints $\mathbf{p}$ from the image. The correspondences ${\mathbf{p}, \mathbf{p}^*}$ feed into the IBVS controller to compute the velocity command for visual servoing.

    \item \textbf{PBVS+PVNet}. To implement PBVS, we estimate the current camera pose $\mathbf{T}_{co}$ per frame using the uncertainty-driven PnP algorithm~\cite{peng2019pvnet} with 2D keypoints from PVNet. The estimated and target poses $\mathbf{T}_{c^*o}$ are used in the PBVS control law~\cite{chaumette2006visual} to compute the camera velocity. Figure~\ref{fig_pbvs} illustrates this process.
\end{itemize}

\subsection{Evaluation Metrics}
Inspired by~\cite{yu2023hyper,felton2023deep}, we use four criteria to quantitatively analyze the servo performance: servo success rate (\textbf{SR}), final translation error (\textbf{TE}), final rotation error (\textbf{RE}), and trajectory length ratio (\textbf{LR}).
\begin{itemize}
\item \textbf{Servo Success Rate (SR)}: A trial is successful if final velocities are near zero and the final pose is accurate. We use the average model distance (ADD) metric, considering a pose accurate if ADD is below 10\% of the object diameter.
\item \textbf{Final Translation Error (TE)}: Defined as the norm of the translation component of the relative transformation $\mathbf{T}_{c^*c}$ between final and desired poses.
\item \textbf{Final Rotation Error (RE)}: The angular difference from the rotation component of $\mathbf{T}_{c^*c}$, computed via axis-angle representation.
\item \textbf{Trajectory Length Ratio (LR)}: Defined as the fraction of the method’s trajectory length divided by the length of the geodesic PBVS trajectory~\cite{felton2023deep}. As illustrated in Figure~\ref{fig_traj}, The geodesic of PBVS is the shortest path in SE(3) connecting the initial and desired camera poses, representing the most efficient trajectory.
\end{itemize}
To ensure fair evaluation, we compute the averages of translation error (\textbf{TE}), rotation error (\textbf{RE}), and trajectory length ratio (\textbf{LR}) only over successful servoing trajectories.

\begin{table*}[t]
\centering
\resizebox{\textwidth}{!}{
\begin{tabular}{lccc ccc}
\toprule
\multirow{2}{*}{Metric} & \multicolumn{3}{c}{Normal Conditions} & \multicolumn{3}{c}{Adverse Conditions} \\
\cmidrule(lr){2-4} \cmidrule(lr){5-7}
& IBVS+PVNet & PBVS+PVNet & Ours+PVNet 
& IBVS+PVNet & PBVS+PVNet & Ours+PVNet \\ 
\midrule
SR (\%)   & 87.81 & 84.15 & \textbf{95.12} & 52.17 & 40.58 & \textbf{82.61} \\
TE (mm)   & 3.27 ± 1.72 & 3.66 ± 1.76 & \textbf{3.17 ± 1.45} & 4.91 ± 4.09 & 5.53 ± 4.08 & \textbf{3.99 ± 2.72} \\
RE (deg)  & 3.91 ± 2.69 & 3.98 ± 3.81 & \textbf{3.81 ± 2.68} & 6.15 ± 3.98 & 6.44 ± 4.16 & \textbf{5.70 ± 3.98} \\
LR (\textbackslash) & 1.25 ± 0.36 & 1.29 ± 0.58 & \textbf{1.11 ± 0.15} & 1.68 ± 0.78 & 1.92 ± 1.02 & \textbf{1.18 ± 0.23} \\
\bottomrule
\end{tabular}
}
\caption{Real-world performance comparison of visual servoing approaches. Bold values indicate the best performance. Errors are reported as mean ± standard deviation over successful trials. Note that, for our approach, the results were obtained without activating the uncertainty-aware velocity reduction, allowing a fair comparison of standard servoing performance.}
\label{tab_results}
\end{table*}

\subsection{Experiment Results}
\begin{figure}[t]
\centering
\begin{subfigure}{0.145\textwidth}
  \includegraphics[width=\linewidth]{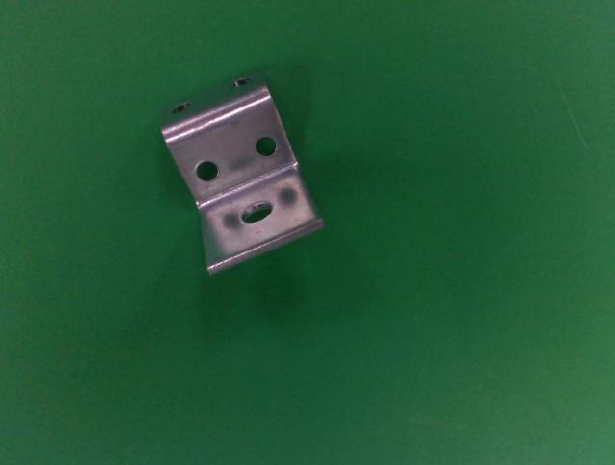}
  \caption{Initial image}
  \label{fig_start}
\end{subfigure}
\begin{subfigure}{0.145\textwidth}
    \includegraphics[width=\linewidth]{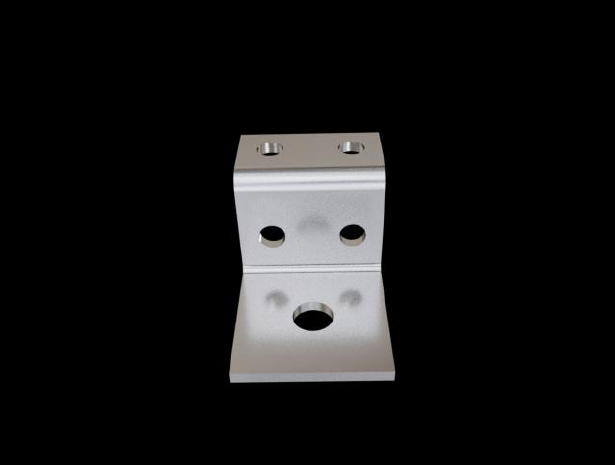}
    \caption{Desired image}
     \label{fig_desired}
\end{subfigure}
\begin{subfigure}{0.145\textwidth}
    \includegraphics[width=\linewidth]{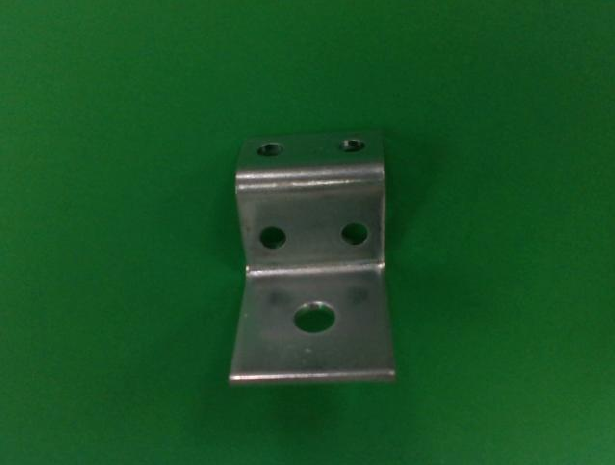}         
    \caption{Final image}
     \label{fig_end}
\end{subfigure}
\begin{subfigure}{0.235\textwidth}
    \includegraphics[width=\linewidth]{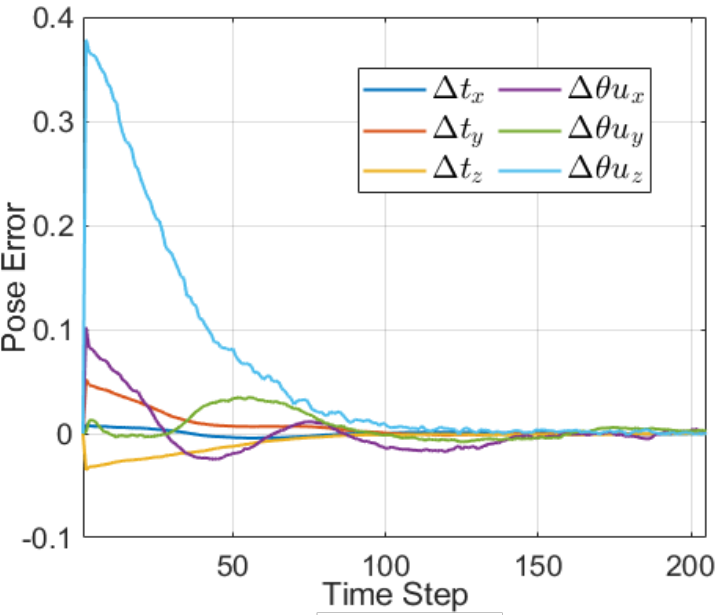}         
    \caption{Pose error}
     \label{fig_pose}
\end{subfigure}
\begin{subfigure}{0.235\textwidth}
    \includegraphics[width=\linewidth]{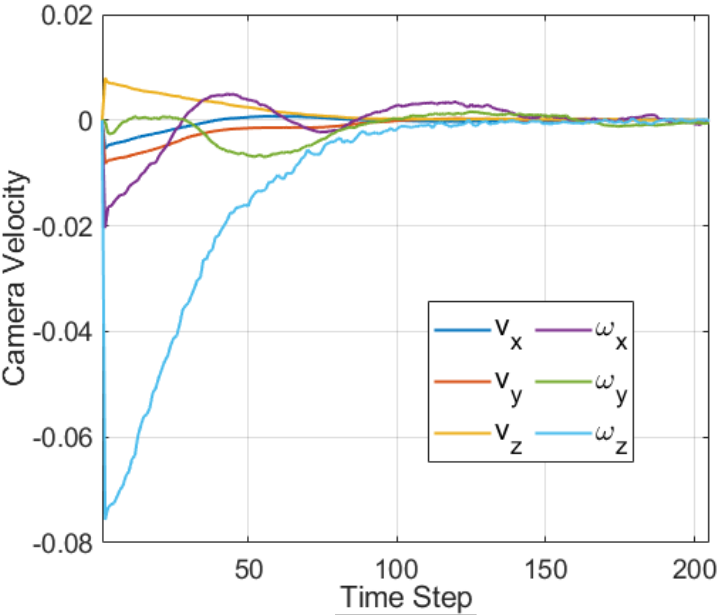}         
    \caption{Camera velocity}
     \label{fig_vel}
\end{subfigure}
\begin{subfigure}{0.28\textwidth}    
   \includegraphics[width=\linewidth]{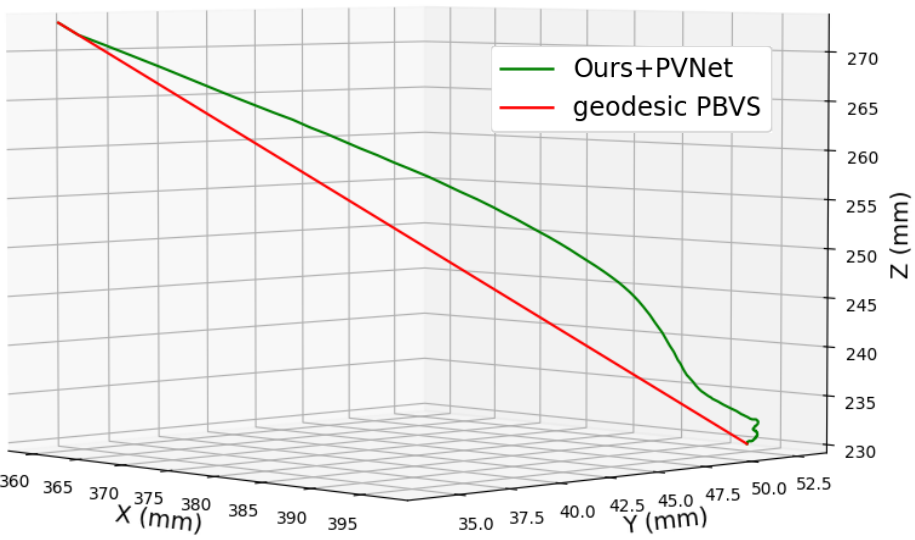}         
   \caption{3D trajectories}
    \label{fig_traj}
\end{subfigure}
\caption{Example visual servoing experiment on the object "Zigzag": (a) Initial image. (b) Desired image (rendered with desired pose). (c) Final image after servoing. (d) Pose error over time (units: meters and radians). (e) Camera velocity commands (units: meters and radians). (f) 3D trajectories of our approach and geodesic PBVS.}
\label{fig_plots}
\end{figure}

We first present an example of our perception-control coupled visual servoing in Figure~\ref{fig_plots}. The initial and desired image, depicted in Figures~\ref{fig_start} and \ref{fig_desired}, reveal a significant initial pose discrepancy. As the servoing progresses, our method rapidly reduces the pose error and ensures a smooth decrease in camera velocity. Ultimately, the pose error is minimized, and the camera velocity converges to zero, demonstrating the effectiveness of our method. This behavior results in a stable and smooth camera trajectory (Figure~\ref{fig_traj}) that closely resembles the geodesic path of a PBVS strategy. 

Table~\ref{tab_results} presents the servoing performance of different approaches evaluated with multiple trials. Our method consistently outperforms both IBVS+PVNet and PBVS+PVNet in success rate (\textbf{SR}), achieving \textbf{95.12\%} in normal conditions and \textbf{82.61\%} in adverse conditions, demonstrating superior robustness. Moreover, our approach produces the lowest translation error (\textbf{TE}), rotation error (\textbf{RE}), and length ratio (\textbf{LR}) in both scenarios, reducing these metrics by up to 0.92 mm, 0.45 degrees, and 0.5, respectively. These findings highlight the advantages of our method in enhancing accuracy, efficiency, and reliability for visual servoing with textureless objects in real-world environments.

\begin{figure}[t]
\centering
\begin{subfigure}{0.235\textwidth}
  \includegraphics[width=\linewidth]{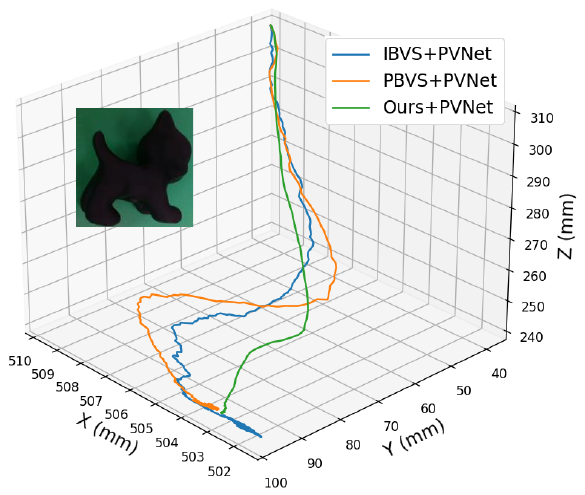}
  \caption{Object "Cat"}
  \label{traj_cat}
\end{subfigure}
\begin{subfigure}{0.235\textwidth}
    \includegraphics[width=\linewidth]{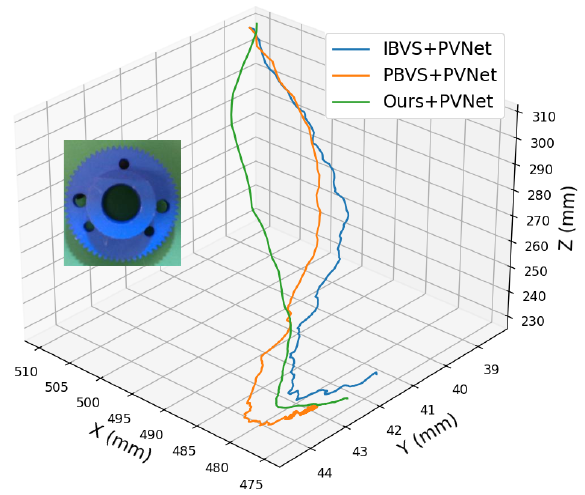}
    \caption{Object "Gear"}
     \label{traj_gear}
\end{subfigure}
\begin{subfigure}{0.235\textwidth}
   \includegraphics[width=\linewidth]{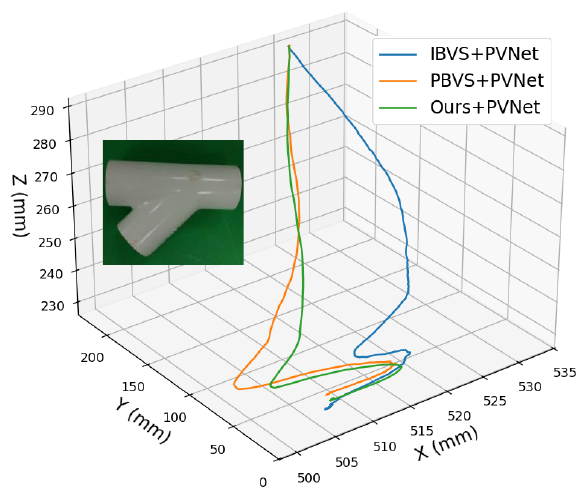}         
   \caption{Object "Pipe"}
    \label{traj_pipe}
\end{subfigure}
\begin{subfigure}{0.235\textwidth}
  \includegraphics[width=\linewidth]{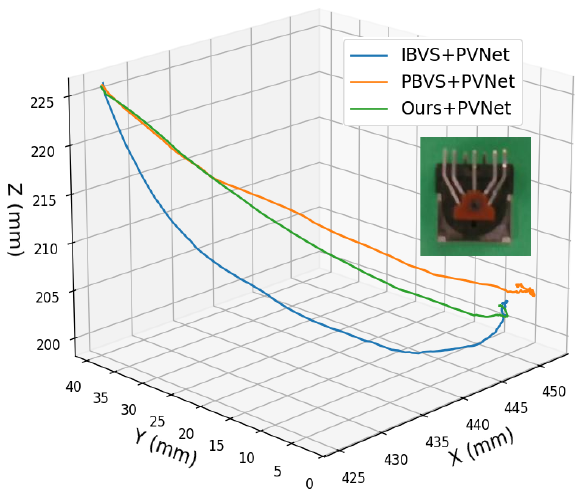}         
   \caption{Object "Connector"}
    \label{traj_sdf}
\end{subfigure}
\caption{Comparison of 3D trajectories during visual servoing on different objects using IBVS, PBVS, and our method.}
\label{fig_all_trajs}
\end{figure}

Figure~\ref{fig_all_trajs} further compares the 3D trajectories of our approach against baseline methods, using the same PVNet measurements. Compared to IBVS and PBVS, the trajectories generated by our method are shorter and smoother, further demonstrating its effectiveness and stability.

\subsection{Uncertainty Evaluation}
\begin{figure}[t]
\centering
  \includegraphics[width=0.97\linewidth]{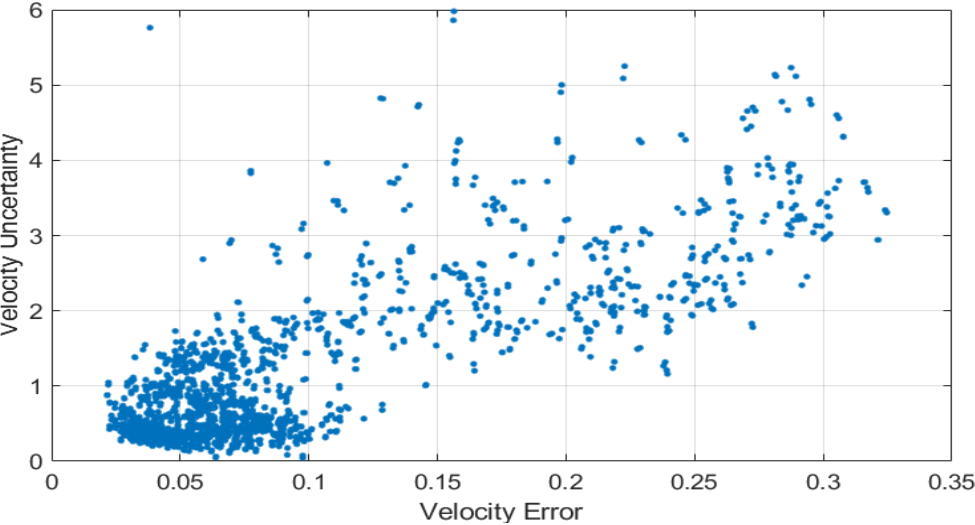}
\caption{The predicted velocity uncertainty correlates well with the velocity error. We compute the Pearson correlation coefficient (0.81 in this example).}  
\label{fig_unc}
\end{figure}

To evaluate the estimated velocity uncertainties, we compared the error in our estimated velocities against their corresponding uncertainty estimates. For each frame, we manually labeled the object pose, $\mathbf{T}_{co}^{GT}$, and computed the ground truth camera velocity, $\mathbf{v}_{c}^{GT}$, using the labeled pose, $\mathbf{T}_{co}^{GT}$, and the desired camera pose, $\mathbf{T}_{c^*o}$. As illustrated in Figure~\ref{fig_unc}, the estimated uncertainties closely follow the actual velocity errors, indicating that the uncertainty estimates are both reliable and well aligned with our goal.

\begin{table}[b]
\centering
\resizebox{0.49\textwidth}{!}{
\begin{tabular}{lccccc c}
\toprule
Method & Zigzag & Pipe & Gear & Cat & Connector & Avg. \\ 
\midrule
IBVS  & 79.2 & 69.6 & 78.3 & 71.4 & 73.7 & 74.4 \\
PBVS  & 75.0 & 56.5 & 86.9 & 76.2 & 73.7 & 73.7 \\
Ours  & \textbf{95.8} & \textbf{78.3} & \textbf{95.6} & \textbf{90.5} & \textbf{89.5} & \textbf{89.9} \\
\bottomrule
\end{tabular}
}
\caption{Robotic grasping success rate (\%) using visual servoing across different objects.}
\label{tab_grasping}
\end{table}

\subsection{Integration in a Grasping Pipeline}
We further evaluate the applicability of our approach in robotic grasping tasks with textureless objects. In this setup, visual servoing first drives the robot to a target object pose. From this pose, a preprogrammed motion offset, defined as a fixed object-specific rigid transformation, moves the end-effector from the servoing target pose to the grasp pose. The grasp pose is manually specified for each object based on its geometry, and the same object-specific offset is used for all baselines to ensure a fair evaluation.

For each object, we conduct 19–23 servoing trials using both our method and baseline approaches. The lighting conditions, as well as the desired and initial camera poses, match those in Section~\ref{setup}, with the initial poses sampled from a broader range of rotations up to $\left[-105^{\circ}, 105^{\circ}\right]$.

Table~\ref{tab_grasping} summarizes the grasping results. Our approach consistently outperforms both IBVS and PBVS for all objects, achieving the highest average success rate of 89.9\%. These results demonstrate the effectiveness and robustness of our method in real-world robotic grasping scenarios.

\section{CONCLUSION}
In this work, we proposed a perception–control coupled visual servoing framework that operates solely on RGB images, enabling robust servoing of textureless objects while maintaining performance under adverse conditions. By integrating per-frame object keypoints with motion priors through an Extended Kalman Filter (EKF), the framework achieves accurate and stable 6D pose estimation for effective pose-based visual servoing. We further introduced a probabilistic control law that models velocity command uncertainties, enhancing the safety of robotic operations. Experimental results on real-world robotic systems show that the proposed framework improves the performance of classical visual servoing algorithms, achieving higher accuracy and robustness under diverse conditions.

Future directions include extending the framework to dynamic environments, incorporating active viewpoint selection, and generalizing to more challenging object types such as CAD-less or deformable objects. For example, high uncertainty in the estimated velocity could trigger active perception, enabling motion replanning to better handle occlusions or dynamic obstacles.

\section*{Acknowledgements}
This work was developed with the assistance of OpenAI’s GPT-4, which was used to (i) generate and refine code for robotic system integration, and (ii) editing and grammar enhancement for the manuscript. All AI-generated content was carefully reviewed and validated by the authors.






\bibliographystyle{ieeetr}
\bibliography{bibliography.bib}

\end{document}